\documentclass{article}

\PassOptionsToPackage{numbers,sort,compress}{natbib}

\usepackage[nonatbib, main, final]{neurips_2026}

\usepackage[
    backend=biber,
    style=numeric-comp,
    natbib=true,
    sorting=nyt,
    maxbibnames=99
]{biblatex}
\AtEveryBibitem{%
  \clearfield{month}
  \clearfield{isbn}
  \clearfield{issn}
  \clearfield{url}
  \clearfield{urldate}
  \clearfield{note}
  \clearfield{urldate}
  \clearfield{pages}
}
\AtEveryBibitem{\clearfield{urldate}}
\DeclareFieldFormat{urldate}{}

\usepackage[utf8]{inputenc} %
\usepackage[T1]{fontenc}    %
\usepackage{url}            %
\usepackage{booktabs}       %
\usepackage{amsfonts}       %
\usepackage{nicefrac}       %
\usepackage{microtype}      %
\usepackage[dvipsnames]{xcolor}      %
\usepackage[
    colorlinks=true,
    linkcolor=BrickRed,
    citecolor=BrickRed,
    urlcolor=BrickRed
]{hyperref}

\usepackage{amsmath}
\usepackage{graphicx}
\usepackage{algorithm}
\usepackage{algpseudocode}
\usepackage[skip=5pt]{caption}
\usepackage{placeins}

\usepackage{xspace}
\usepackage[inline]{enumitem}

\usepackage{wrapfig}

\newcommand{\cf}{cf.\xspace}

\title{RAIL: Representation-Aligned Imitation Learning \\ for Student-Compatible Teacher Policies}

\author{%
  Meraj  Mammadov \\
  Örebro University\\
  \And
  Pedro Zuidberg Dos Martires \\
  Örebro University\\
  \And
  Johannes A. Stork \\
  Örebro University\\
}

\begin{document}
\maketitle

\begin{abstract}
Reinforcement learning (RL) from raw sensory inputs such as images or onboard sensors can be challenging due to high-dimensional observations and sparse rewards.
A common strategy is to train a teacher policy with access to privileged state information and then distill it into a student policy that acts from raw inputs alone.
However, when the teacher relies on information unavailable to the student, exact imitation may be impossible, creating an irreducible imitation gap.
Existing approaches typically mitigate this mismatch through careful reward shaping or additional reinforcement learning on the student policy after the teacher has been trained. 
Instead, we introduce Representation-Aligned Imitation Learning (RAIL), which addresses the mismatch during teacher training itself.
RAIL learns a latent representation shared across teacher and student observations using contrastive learning, and trains the teacher policy directly in this space.
Because the teacher policy is constrained to operate on this shared space, it learns behaviors that are reproducible from student observations.
This substantially reduces the imitation gap while preserving task performance.
Across multiple environments, RAIL outperforms strong baselines without reward modification or post-hoc student fine-tuning, and surpasses direct reinforcement learning from raw observations. 
The learned representations further enable zero-shot transfer to new tasks through teacher-only training. 
Project site: \href{https://merajmammadov.com/rail/}{https://merajmammadov.com/rail/}.

\end{abstract}

\section{Introduction}
Training reinforcement learning policies directly on the target observation space can be impractical due to high dimensional observation space, sparse rewards, or costly exploration. A common strategy is therefore to first train a teacher policy with RL using privileged state information, and then to distill its behavior into a student policy that operates with the actual deployment observations~\citep{chen2019learning}. For example, in navigation tasks the teacher may observe a full map, obstacle locations, or target coordinates, while the student must act on onboard camera images or limited local sensing.

\begin{wrapfigure}[13]{r}{0.25\textwidth}
    \centering
    \includegraphics[width=0.25\textwidth]{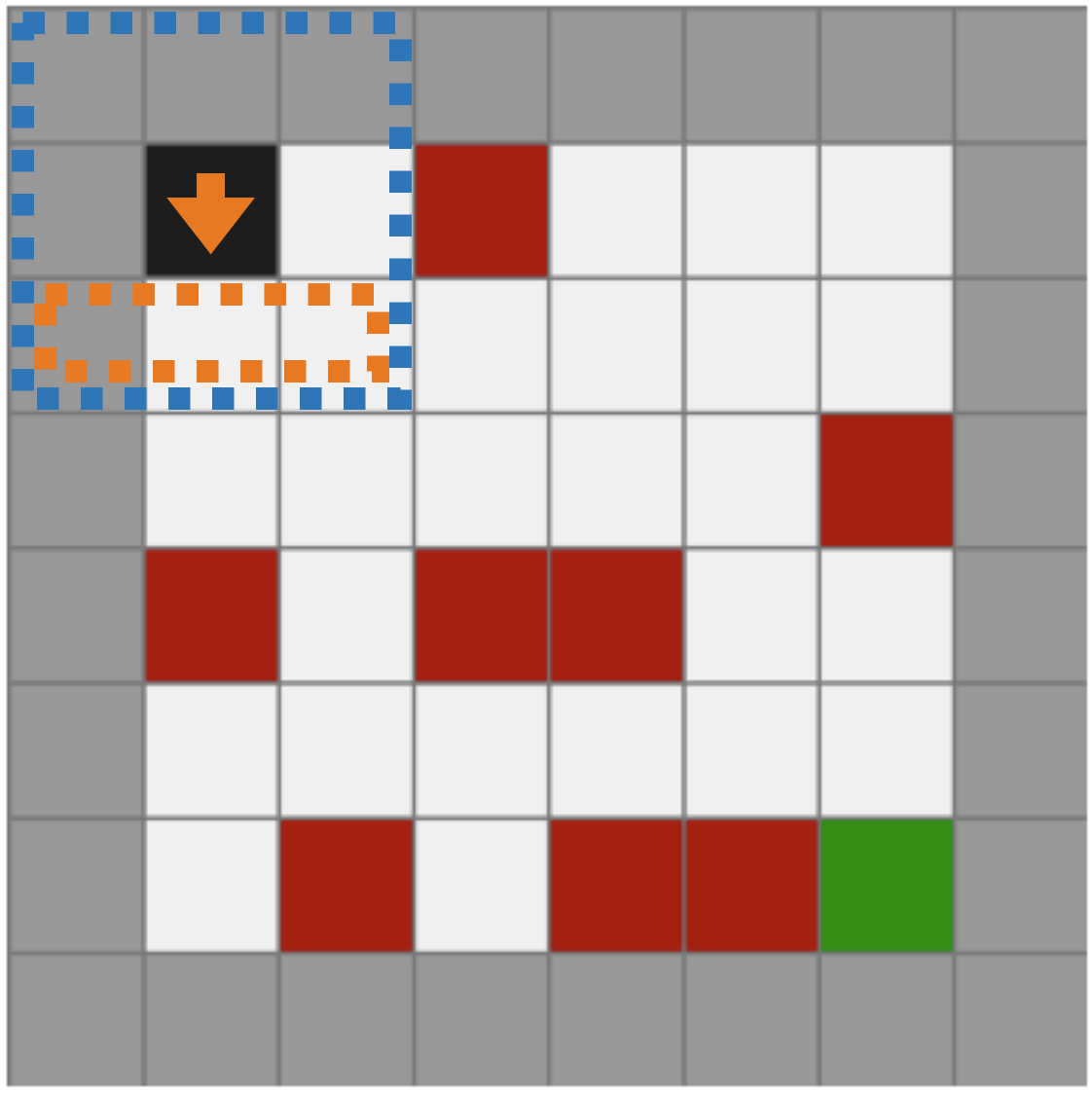}
    \caption{Observation mismatch in TunnelVision: Teacher (blue) vs.\ student (orange).}
    \label{fig:intro_example}
\end{wrapfigure}

While this teacher--student paradigm can greatly simplify training, it also introduces a fundamental mismatch:
when the teacher exploits privileged variables that are not recoverable from the student's observations, the resulting policy may be impossible for the student to reproduce. In that case, no policy conditioned only on student observations can exactly match the teacher's decisions, leading to an irreducible imitation gap~\citep{walsman2023impossibly}.

To illustrate this, consider the navigation task to the right (Figure \ref{fig:intro_example}), where an agent is dropped on the black tile and must reach the goal (green tile), while avoiding obstacles (red tiles). A privileged teacher observes all neighboring cells (dashed blue) and can move toward the goal without the need to change orientation. In contrast, the student observes only three cells in front of it (dashed orange) and needs to actively change its orientation to complete the mission successfully.
This means that the teacher can distinguish between states that are indistinguishable to the student, which in turn means that the teacher is capable of taking more informed actions than the student. 

Existing approaches typically address this mismatch after teacher training, either by incorporating teacher signals into the student's RL objective \citep{mosbach2024grasp, nguyen2022leveraging}, or by additional RL fine-tuning of the student \citep{xing2024bootstrapping}. While effective in certain settings, these methods reintroduce the cost of RL optimization and do not remove the underlying information asymmetry.

In this work, we address the teacher--student imitation gap at its source: instead of training an unconstrained teacher and repairing the student afterward, we constrain the teacher to act only on information accessible to the student. We propose \emph{Representation-Aligned Imitation Learning} (RAIL), which learns a common representation of teacher and student observations that captures only the information available to both. The teacher policy is trained with RL in this space, and the same policy is applied by the student at deployment time.

The contributions of this paper are:
\begin{enumerate*}[label=(\arabic*)]
    \item We propose RAIL, a teacher--student imitation framework that enforces imitability during teacher training by restricting the teacher policy to a shared latent space that is also recoverable from student observations.

    \item We introduce a contrastive representation-learning objective with gradient blocking, alignment, and stability regularization to prevent the leakage of private teacher information and enable direct policy transfer.

    \item Across multiple environments, we show that RAIL reduces the imitation gap, outperforms strong baselines, learns faster than pixel-based RL, and supports zero-shot transfer to new tasks.
\end{enumerate*}

\section{Preliminaries}
\label{sec:preliminaries}
We consider a sequential decision-making problem in which a teacher $T$ and a student $S$ act in the same environment but observe it through different observation mappings. Concretely, the environment is defined by the state space $\mathcal S$, action space $\mathcal A$, transition kernel $P(\cdot \mid s, a)$, and reward function $r : \mathcal S \times \mathcal A \to \mathbb R$. At time-step $t$, both agents observe the same latent state $s_t \in \mathcal S$ through their corresponding observation mappings, $o_t^T = f_T(s_t)$ and $o_t^S = f_S(s_t)$, where $o^T_t \in \mathcal O_T$ and $o^S_t \in \mathcal O_S$ denote teacher and student observations, respectively.
The teacher and student follow policies $a_t^T \sim \pi_T(\cdot \mid o_t^T)$ and $a_t^S \sim \pi_S(\cdot \mid o_t^S)$ that map their respective observations to a probability distribution over their shared action space.

In classical teacher--student settings \citep{chen2019learning, nguyen2022leveraging}, the teacher policy $\pi_T$ is trained with reinforcement learning to maximize expected discounted return:
\begin{equation}
\max_{\pi_T} \;
\mathbb E_{\tau \sim \pi_T}
\left[
\sum_{t=0}^\infty \gamma^t r(s_t, a_t^T)
\right],
\end{equation}
where $\tau = (s_0,a_0,\dots)$ is a trajectory generated by $\pi_T$ under the transition kernel $P$ and $\gamma \in (0, 1]$ is the discount factor. The student policy $\pi_S$ is later trained to approximate the teacher's policy under its own observation space via imitation learning.

Given paired observations $(o^T_t,o^S_t)$ from the same state $s_t$, standard behavior cloning \citep{pomerleau1989alvinn} fits a student policy $\pi_S$ to match a fixed teacher policy $\pi_T$ by minimizing a divergence $D(\cdot,\cdot)$ between the action distribution of the teacher and student policies, conditioned on their respective aligned observations:
\begin{equation}
\min_{\pi_S}\;
\mathbb{E}_{(o^T_t,o^S_t)\sim\mathcal D}
\!\left[
D\!\left(
\pi_T(\cdot \mid o^T_t),\,
\pi_S(\cdot \mid o^S_t)
\right)
\right],
\label{eq:imitation}
\end{equation}
where $\mathcal D = \{ (o_t^T, o_t^S)\}_t$ is a dataset of aligned observations collected from trajectories $\tau \sim \pi_T$. 

\subsection{Irreducible imitation gap}
\label{subsec:imitation_gap}
Behavior cloning implicitly assumes that the teacher policy is realizable under the student's observation space. When this assumption holds, small imitation error translates to small performance gap between the agents. In particular, if the policy mismatch in Eq.~\eqref{eq:imitation} is upper-bounded by $\varepsilon$, and rewards are bounded as $|r(s,a)| \le R_{\max}$, the return gap satisfies the bound of \citet{xu2020error}:
\begin{equation}
V_{\pi_T}-V_{\pi_S}
\le
\frac{2\sqrt{2}\,R_{\max}}{(1-\gamma)^2}\sqrt{\varepsilon},
\label{eq:imitation_gap}
\end{equation}
where $V_\pi$ denotes expected discounted return. Thus, even small one-step imitation error can induce substantial long-horizon performance loss.

A sufficient condition for zero imitation gap is that the teacher action be conditionally independent of teacher observation given the student observation,
\begin{equation}
a^T \perp o^T \mid o^S.
\label{eq:conditional_independence}
\end{equation}
That is, once $o^S$ is known, any additional variation in the teacher's observation does not affect its action distribution. This implies that the teacher policy is a function of $o^S$ alone and can therefore be matched exactly by a student policy conditioned on $o^S$.

In practice, however, the condition in Eq.~\eqref{eq:conditional_independence} is often violated because the teacher bases its decisions on information unavailable to the student. Then there may exist two distinct states $s_1,s_2 \in \mathcal S$ that are identical to the student:
\begin{equation}
f_S(s_1)=f_S(s_2),
\end{equation}
while the teacher distinguishes them and acts differently:
\begin{equation}
\pi_T(\cdot \mid f_T(s_1)) \neq \pi_T(\cdot \mid f_T(s_2)).
\end{equation}
Because the student conditions only on $o^S=f_S(s_1)=f_S(s_2)$, it assigns the same action distribution to both states and therefore cannot match the teacher when distinguishing the states is necessary. The minimum achievable imitation loss is therefore strictly positive whenever such indistinguishable states occur with nonzero probability.

As shown by \citet{ross2011reduction}, these local mismatches can compound over time, and we also show this in our experimental section (\cf Section~\ref{sec:experiments}).
In particular, we show that such errors drive the student off-distribution, prevent recovery, and ultimately lead to large teacher--student performance gaps in state-of-the-art imitation learning approaches~\citep{pomerleau1989alvinn, ross2011reduction}, while RAIL successfully mitigates the presence of such scenarios and achieves a low imitation gap and high student performance.

\section{Representation-Aligned Imitation Learning}
\label{sec:method}
\begin{figure}[t]
\centering
\includegraphics[width=1.0\textwidth]{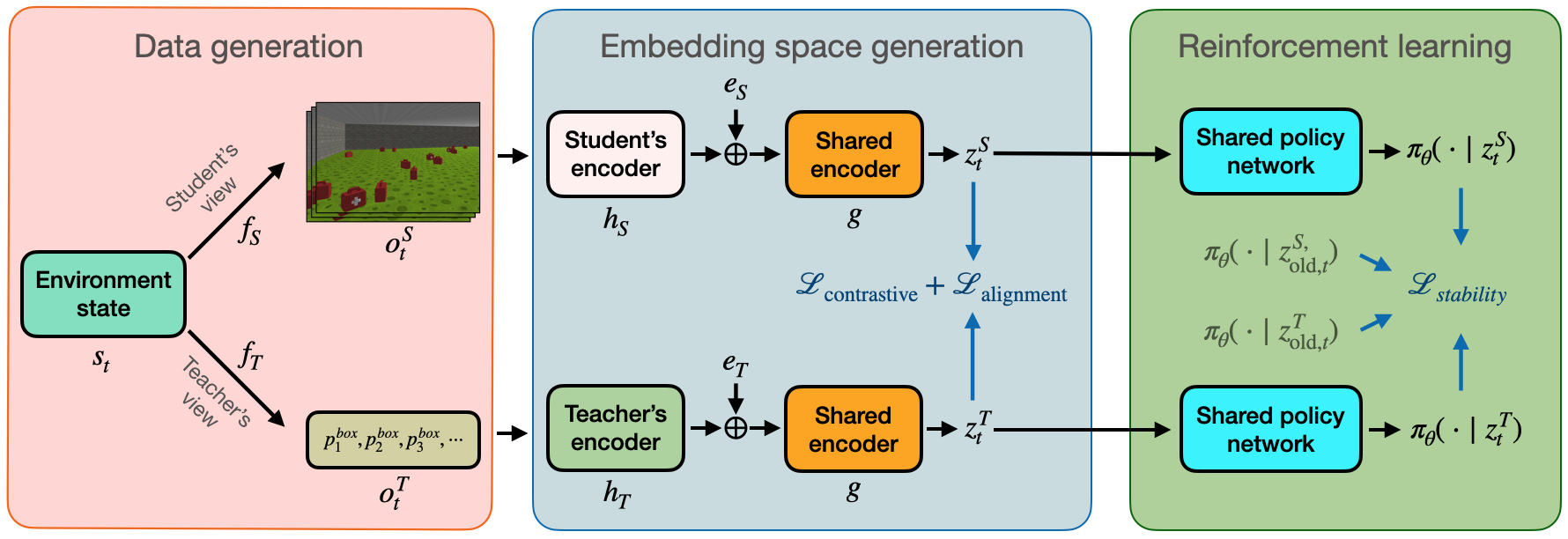}
\caption{Left: teacher and student observations are generated from the same underlying state via different observation mappings. 
Middle: teacher and student encoders $h_T$ and $h_S$, followed by a shared encoder $g$, map both observations into a common latent space; $e_T$ and $e_S$ are learned agent identity embeddings.
Right: a shared policy $\pi_\theta$ operates on these representations to produce action distributions $\pi_\theta(\cdot \mid z^{T}_t)$ and $\pi_\theta(\cdot \mid z^{S}_t)$.
$z^{T}_{\mathrm{old},t}$ and $z^{S}_{\mathrm{old}, t}$ denote embeddings computed using a frozen copy of the encoders and are used to evaluate the stability objective $L_{\mathrm{stability}}$ (\cf Subsection~\ref{subsec:stability}).
Blue arrows indicate the losses used to train the encoders (\cf Section~\ref{sec:method}).
}
\label{fig:overview}
\end{figure}

RAIL enforces imitability at the level of the teacher policy by restricting it to depend only on information recoverable from the student observations. Instead of training an unconstrained teacher and compensating for the resulting mismatch afterward, we constrain the policy input during training so that all decisions are realizable under the student observation space. 

RAIL builds on the view that teacher and student observations contain overlapping information about the same underlying state, together with private information available only to one side \citep{correia2022multiview}. Exact transfer requires the teacher policy to depend only on the overlapping information. We therefore learn encoders $\phi_T$ and $\phi_S$ that map teacher and student observations into a common latent space:
\begin{equation}
    z^T_t = \phi_T(o^T_t), \qquad z^S_t = \phi_S(o^S_t).
\end{equation}
For aligned observations from the same state $s_t$, we encourage $z^T_t$ and $z^S_t$ to agree while keeping representations of different states distinguishable. This naturally leads to a contrastive objective: paired teacher--student observations are treated as positives, while observations from different states act as negatives. Optimizing this objective encourages the latent space to retain information shared across both observation spaces while suppressing factors specific to each observation.

We train a single policy $\pi_\theta$ on top of this learned representation. During training, the teacher acts through teacher embeddings,
\begin{equation}
a \sim \pi_\theta(\cdot \mid z^T),
\end{equation}
and at deployment the student applies the same policy to student embeddings,
\begin{equation}
a \sim \pi_\theta(\cdot \mid z^S).
\end{equation}
Because the same policy $\pi_\theta$ operates on both teacher and student embeddings, transfer requires only replacing $\phi_T$ with $\phi_S$ once the embeddings are aligned.

An overview of the framework is shown in Figure~\ref{fig:overview}. RAIL iteratively updates the encoder networks $\phi_T$ and $\phi_S$ with contrastive learning and the policy network $\pi_\theta$ with reinforcement learning on the learned embeddings. We describe this in more detail in Subsection~\ref{sec:algorithm}.

To learn imitable teacher policies that allow direct transfer to the student, RAIL jointly optimizes three objectives: a contrastive loss to construct the shared latent space (\cf Subsection~\ref{sec:loss:contrastive}), an alignment loss to enable direct policy transfer (\cf Subsection~\ref{sec:loss:alignment}), and a stability loss to regularize representation updates during policy learning (\cf Subsection~\ref{subsec:stability}).

\subsection{Learning a shared latent space}
\label{sec:loss:contrastive}
We learn the shared latent space from aligned teacher--student observation pairs collected during teacher rollouts. For each pair $(o_t^T,o_t^S)$ corresponding to the same state $s_t$, the embeddings $z_t^T$ and $z_t^S$ should agree, while embeddings from different states should remain distinguishable. 
We use a contrastive objective~\cite{oord2018representation} as its positive-negative structure directly matches this goal.

We parameterize the encoders as $\phi_T = g \circ h_T$ and $\phi_S = g \circ h_S$, where $h_T$ and $h_S$ map teacher and student observations into feature spaces, and $g$ maps both into a shared latent space. Given an observation $o_i^A$ from agent $A \in \{T,S\}$, we compute
\begin{equation}
    z_i^A = \phi_A(o_i^A) = \frac{g(h_A(o_i^A), e_A)}{\|g(h_A(o_i^A), e_A)\|_2},
\end{equation}
where $e_A \in \mathbb R^4$ is a learned indicator of the input source.

The contrastive loss over the embeddings is given by the symmetric InfoNCE objective~\citep{oord2018representation}:
\begin{equation}
\mathcal{L}_{\mathrm{contrastive}}
=
-\frac{1}{2N}\sum_{i=1}^{N}
\left[
\log
\frac{\exp((z_i^T)^\top z_i^S/\tau)}
{\sum_{j=1}^{N}\exp((z_i^T)^\top z_j^S/\tau)}
+
\log
\frac{\exp((z_i^S)^\top z_i^T/\tau)}
{\sum_{j=1}^{N}\exp((z_i^S)^\top z_j^T/\tau)}
\right],
\label{eq:contrastive}
\end{equation}
where $z_i^T$ and $z_i^S$ are unit-normalized embeddings of aligned observations, $N$ is the batch size, and $\tau$ is a learnable temperature parameter.

For each embedding, the aligned embedding from the other agent is treated as positive, while all other embeddings from that agent in the batch serve as negatives. This encourages agreement only on information shared between the teacher and student observations and suppresses factors specific to each~\cite{vonkugelgen2021contentstyle}. 

\subsection{Aligning learned embeddings}
\label{sec:loss:alignment}
The shared policy $\pi_\theta$ is trained on teacher embeddings but deployed on student embeddings. To enable this direct policy transfer, paired teacher--student observations must therefore map to nearly identical latent embeddings. The contrastive objective (Eq.~\eqref{eq:contrastive}) enforces relative similarity by distinguishing aligned pairs from negatives, but it does not minimize the distance between positive pairs directly. Therefore, small residual differences can remain and translate into an imitation gap, especially as reinforcement learning policies can be highly sensitive to input perturbations~\citep{jesse2018generalization}.
To address this, we introduce an absolute pairwise alignment objective that directly maximizes the cosine similarity between embeddings of paired observations:
\begin{equation}
    \mathcal L_\text{alignment}
    =
    -\frac{1}{N} \sum_{i=1}^{N}
    \frac{(z_i^T)^\top z_i^S}
    {\lVert z_i^T \rVert_2\,\lVert z_i^S \rVert_2}
    .
\end{equation}
Empirically, we find that this term is critical for reducing the imitation gap and improving transfer performance (\cf Section~\ref{sec:ablation}).

\subsection{Stabilizing policy learning}
\label{subsec:stability}
In RAIL, the encoder parameters are updated throughout training, which induces a non-stationary input distribution for the policy. As the embedding space evolves, the same observation may map to different latent representations, resulting in unstable policy outputs. 
Moreover, contrastive representations may continue to drift even after alignment appears saturated \citep{betser2026infonce} and further increase this non-stationarity.

We therefore regularize encoder updates to preserve the current policy behavior. Specifically, we penalize changes in the policy outputs induced by representation updates:
\begin{equation}
    \mathcal L_{\text{stability}} = 
    -\frac{1}{N} \sum_{i=1}^{N} 
    \left(
    \frac{(u^{T}_{\text{old}, i})^\top u^T_i}
    {\lVert u^{T}_{\text{old}, i} \rVert_2\,\lVert u^T_i \rVert_2}
    +
    \frac{(u^{S}_{\text{old}, i})^\top u^S_i}
    {\lVert u^{S}_{\text{old}, i} \rVert_2\,\lVert u^S_i \rVert_2}
    \right),
\end{equation}
where $u^T_i, u^S_i$ denote the current outputs of the policy network (action logits for discrete policies and distribution parameters for continuous policies), and $u^T_{\text{old},i}, u^S_{\text{old},i}$ are the corresponding outputs for the same observations computed using a copy of the encoder frozen at the start of the encoder updates.

\subsection{Training procedure}
\label{sec:algorithm}

RAIL can be implemented as a standard reinforcement learning loop with an additional representation-learning phase inside each iteration. At iteration $t$, the agent first collects rollouts using the current teacher-side policy, $a \sim \pi_\theta(\cdot \mid \phi_T(o^T))$, exactly as in ordinary policy optimization \citep{schulman2017proximal}. The resulting rollout buffer contains rewards, actions, and paired teacher--student observations $(o^T,o^S)$ from the same underlying states.

Before updating the policy, RAIL performs $K$ encoder updates on minibatches sampled from this buffer. During this inner loop, the policy is held fixed and the encoders are trained to map paired teacher and student observations to a shared latent space. The encoder loss combines the contrastive, alignment, and stability objectives:
\begin{align}
\mathcal{L}_{\mathrm{embed}}
=
\lambda_{\mathrm{con}}\mathcal{L}_{\mathrm{contrastive}}
+
\lambda_{\mathrm{align}}\mathcal{L}_{\mathrm{alignment}}
+
\lambda_{\mathrm{stab}}\mathcal{L}_{\mathrm{stability}}.
\end{align}

The stability term is computed relative to a frozen copy of the encoders taken before the inner loop, so that representation updates preserve the current policy behavior.
Note that in our experimental evaluation (\cf Section~\ref{sec:experiments}) we used 
$
\lambda_{\mathrm{con}}=
\lambda_{\mathrm{align}}=
\lambda_{\mathrm{stab}}=1
$ 
throughout.

After the encoder updates, the encoders are frozen and RAIL performs the usual RL policy update using teacher embeddings $\phi_T(o^T)$ from the rollout buffer.
Crucially, gradients from the RL objective are not propagated into the
encoders. Without this gradient blocking, policy optimization drives the
representation to encode teacher-private information that improves teacher performance but is unrecoverable from student observations, thereby breaking direct transfer.
Thus, RAIL keeps the outer loop identical to standard teacher-side RL while inserting an inner contrastive loop that maintains student-compatible representations.
Algorithm~\ref{alg:rail} summarizes the overall training procedure.

One implementation detail not explicitly shown in the algorithm is the
computation of the stability objective. At the beginning of each
representation-learning phase, we store frozen copies of the current
encoders, denoted $\bar\phi_T$ and $\bar\phi_S$. These reference encoders remain fixed throughout the subsequent $K$ representation updates and define the policy behavior prior to representation refinement. The stability loss penalizes changes in the policy outputs induced by updates to the embeddings relative to these frozen reference encoders.

\begin{algorithm}[t]
\caption{Representation-Aligned Imitation Learning (RAIL)}
\label{alg:rail}
\begin{algorithmic}[1]
\Require Teacher encoder $\phi_T$, student encoder $\phi_S$, policy $\pi_\theta$
\For{each training iteration}
    \State Collect teacher rollouts using $\pi_\theta(\phi_T(o^T))$
    \State Store aligned teacher--student observations
    \For{$K$ iterations  with $\pi_\theta$ fixed}
        \State Update $\phi_T,\phi_S$ with
        $
        \mathcal{L}_{\mathrm{embed}}
        $
    \EndFor
    \State Update policy $\pi_\theta$ with RL while keeping $\phi_T$, $\phi_S$ fixed
\EndFor
\end{algorithmic}
\end{algorithm}

\section{Related Work}
\label{sec:related_work}
Teacher--student setups in imitation learning commonly train a privileged RL teacher with state information and use behavior cloning (BC) \citep{pomerleau1989alvinn} or dataset aggregation (DAgger) \citep{ross2011reduction} to distill it into a deployable student policy operating on restricted or high-dimensional observations \citep{chen2019learning, zhang2021roach}.
This improves teacher learning, but creates a fundamental mismatch: when the teacher uses information unavailable to the student, the student cannot, in general, reproduce its actions, which yields an irreducible imitation gap \citep{weihs2021bridging, walsman2023impossibly, song2026to}. 
In this case, imitation marginalizes over information private to the teacher, leading to suboptimal student policies.

Most prior work treats this mismatch as a \textbf{student-side optimization problem} after the teacher has already been trained. 
Several methods follow imitation with additional reinforcement learning, either by bootstrapping the policy with imitation learning \citep{xing2024bootstrapping}, incorporating teacher guidance into the student's RL objective \citep{mosbach2024grasp, walsman2023impossibly, weihs2021bridging}, or adaptively querying the teacher to improve learning under partial observability \citep{kim2025distilling}. 
COSIL \citep{nguyen2022leveraging} augments the student's RL objective with a soft imitation term that encourages agreement with a privileged teacher.
These methods can recover performance when the student can compensate through exploration, but, as we show in Section~\ref{sec:experiments}, can severely hinder learning when the optimal behavior differs substantially across teacher and student observations.

A second line of work \textbf{enforces imitability during teacher training}. Monaci et al. \citep{monaci2022dipcan} explicitly restrict the teacher observation space to state variables that are known to be recoverable by the student, which requires task-specific prior knowledge. Reward-shaping methods encourage student-compatible behavior through auxiliary objectives \citep{song2023learning,messikommer2026approximate}, but involve careful reward design. SITT \citep{messikommer2025student} jointly trains teacher and student and penalizes teacher behavior that the student cannot match. 
However, it modifies the original task objective and introduces a sensitive environment-dependent trade-off hyperparameter. RAIL instead preserves the original reward function and enforces imitability by training the teacher on a shared latent space that enables direct transfer of the learned policy.

More broadly, our work relates to representation learning methods for transfer under domain shifts. Shon et al. \citep{shon2006learning} construct a common latent space for robotic imitation of human motions. Subsequent work learns representations that are invariant to domain changes using multi-view contrastive objectives \citep{sermanet2018time,correia2022multiview} or adversarial domain confusion \citep{stadie2017third}.
These approaches combine representation learning with policy training to remove nuisance variation (e.g., viewpoint or background), while the same task-related information is present across views.
Under information asymmetry, however, transfer requires excluding information that is not recoverable from the student observations, which we enforce by stopping policy gradients from updating the encoders. Giammarino et al. \citep{giammarino2025visually} train adversarial imitation learning in a contrastive latent space and report failure under this constraint, which highlights the importance of our stability and alignment objectives.

\begin{table*}[t]
\caption{Environments and imitation gaps used in experiments. Further details and the reward functions are provided in Appendix \ref{appx:envs}.}
\centering
\footnotesize
\setlength{\tabcolsep}{5pt}
\renewcommand{\arraystretch}{1.2}

\begin{tabular}{p{0.065\textwidth} p{0.20\textwidth} p{0.20\textwidth} p{0.20\textwidth} p{0.20\textwidth}}
\toprule

&
\centering
\includegraphics[width=0.15\textwidth]{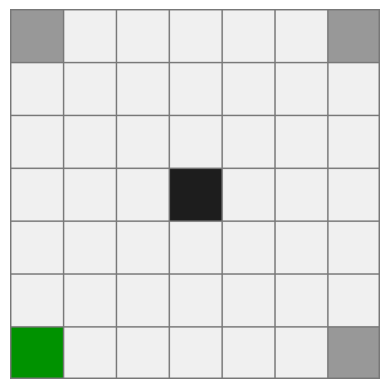}
&
\centering
\includegraphics[width=0.15\textwidth]{figures/tunnel_vision_3.png}
&
\centering
\includegraphics[width=0.15\textwidth]{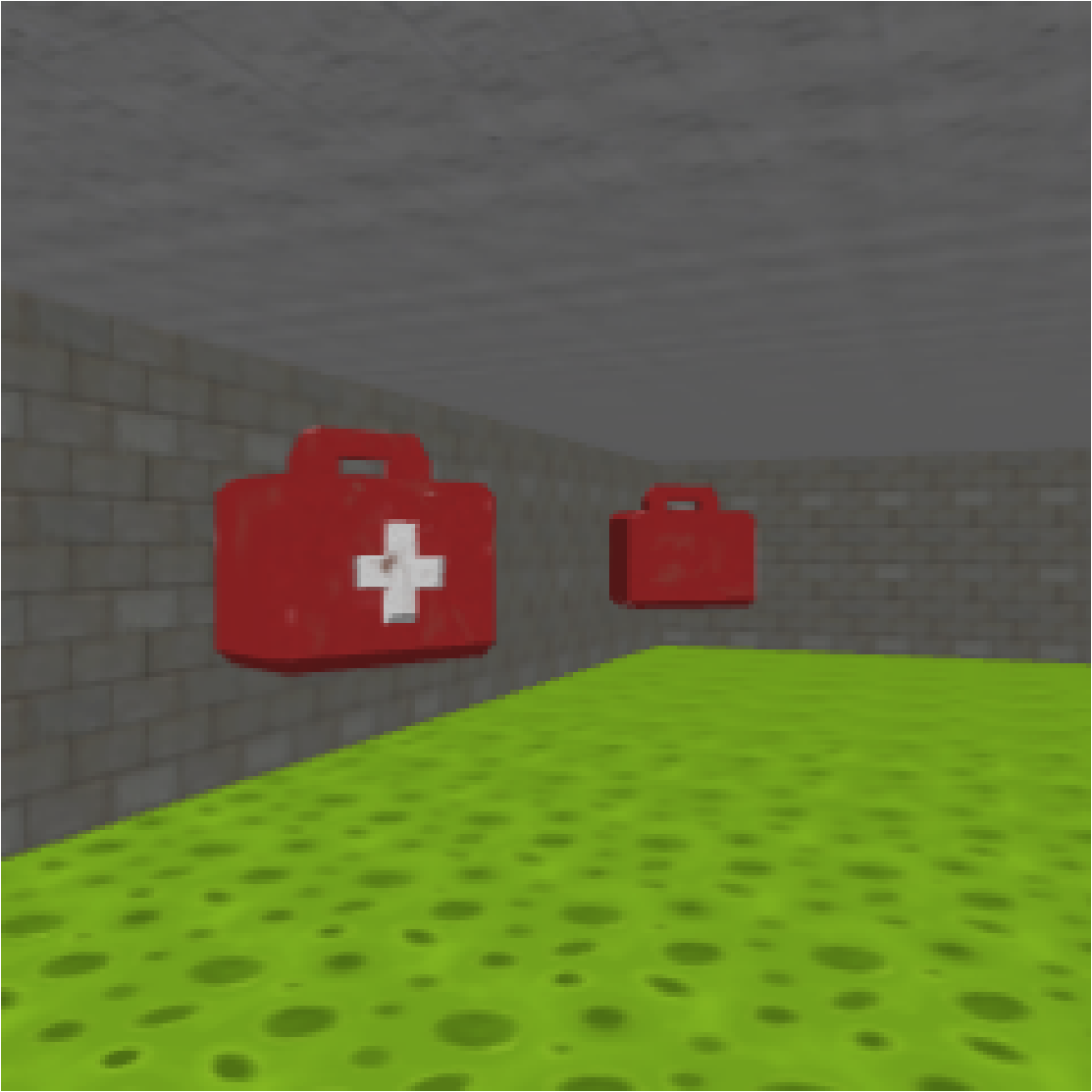}
&
\centering
\includegraphics[width=0.15\textwidth]{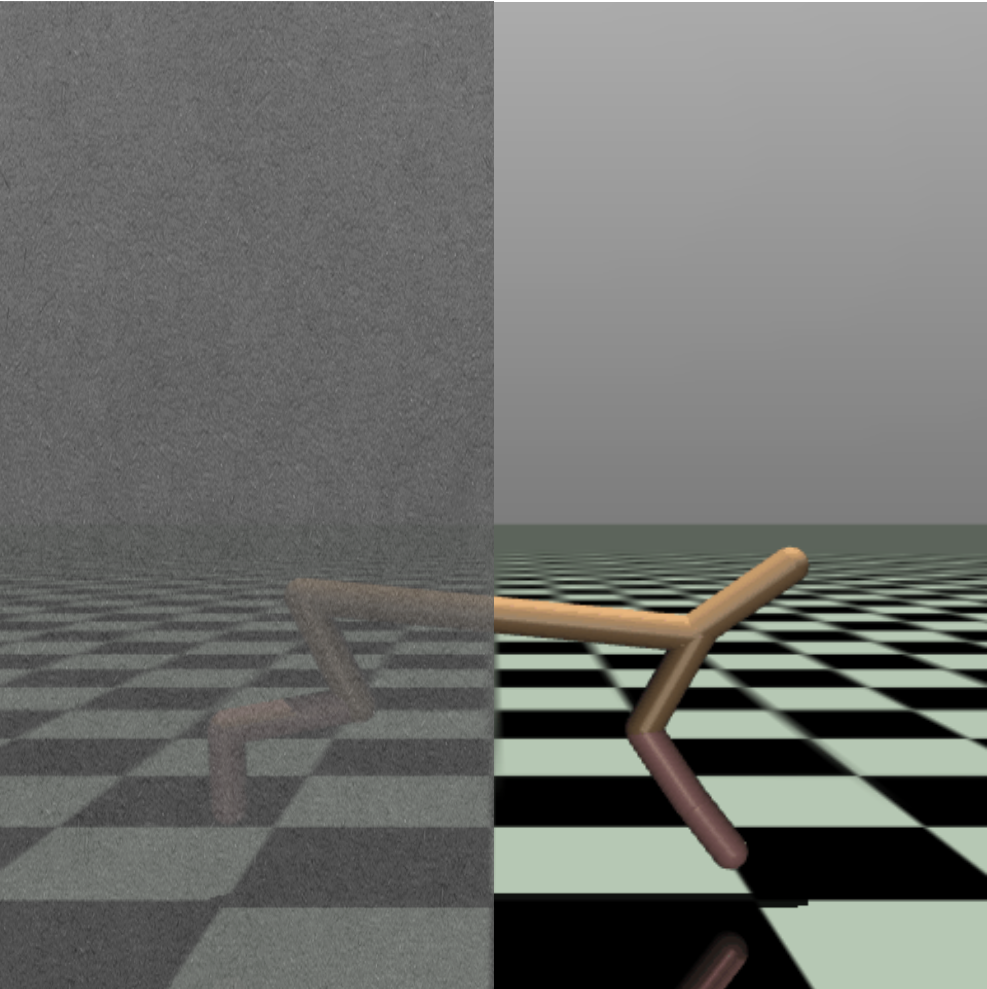}
\tabularnewline

&
\centering \textbf{Corners}
&
\centering \textbf{TunnelVision}
&
\centering \textbf{CollectHealth}
&
\centering \textbf{QuarterCheetah}
\tabularnewline

\midrule

\textbf{Task}
&
Navigate from center to the hidden target corner (green).
&
Reach the goal (green) while avoiding obstacles (red).
&
Collect all health boxes in the room while avoiding walls.
&
Run forward as fast as possible.
\tabularnewline

\midrule

\parbox[t]{\linewidth}{\textbf{Teacher}\\\textbf{obs.}}
&
Position, previous action, and the target corner identity.
&
Position, previous action, and occupancy of all neighboring cells (blue).
&
Position, previous action, and relative coordinates of all remaining boxes.
&
Full proprioceptive state (17D joint positions and velocities).
\tabularnewline

\midrule

\parbox[t]{\linewidth}{\textbf{Student}\\\textbf{obs.}}
&
Position and previous action only; target identity is hidden.
&
Position, previous action, and three neighboring cells facing its current orientation (orange).
&
Stack of 5 limited field-of-view first-person RGB observations.
&
Third-person RGB observation with partial occlusion (left half is removed).
\tabularnewline

\midrule

\parbox[t]{\linewidth}{\textbf{Action}\\\textbf{space}}
&
Move to any neighboring cells, $|\mathcal A|=8$.
&
Move or rotate to compass directions, $|\mathcal A|=8$.
&
Move in 8 directions or rotate $\pm 30^\circ$, $|\mathcal A|=10$.
&
Continuous torques, $|\mathcal A|=6$.
\tabularnewline

\midrule

\parbox[t]{\linewidth}{\textbf{Imitation}\\\textbf{gap}}
&
Teacher can move directly to the target, student must visit all corners.
&
Teacher can move directly to goal, student must turn to discover obstacles.
&
Teacher can move without turning, student must turn to discover next target.
&
Teacher has full state; student must infer dynamics and state from partial visual observations.
\tabularnewline

\bottomrule
\end{tabular}
\label{tab:envs}
\end{table*}

\FloatBarrier
\section{Experiments}
\label{sec:experiments}
We evaluate RAIL in simulation environments with asymmetric observation spaces that induce an irreducible imitation gap. The teacher has access to privileged information, whereas the student operates under restricted observations, which makes direct imitation of an unconstrained teacher insufficient. Table~\ref{tab:envs} summarizes the environments; full details are provided in Appendix~\ref{appx:envs}. 
Our experiments address four questions: (i) whether RAIL reduces the imitation gap, (ii) how it compares to strong baselines, (iii) the contribution of its components, and (iv) whether the learned representation can be used to zero-shot transfer new behaviors through teacher-only training.

\subsection{\texorpdfstring{Experimental setup}{Experimental setup}}
\label{sec:baselines}

We compare RAIL against four imitation learning baselines, along with direct RL from student observations (\textbf{Student RL}) as a reference and which requires reward signals in the student domain.

\textbf{(1) Behavior Cloning (BC) \citep{pomerleau1989alvinn}} 
We train the teacher with RL, and the student is trained offline to imitate teacher actions via supervised learning on collected trajectories.

\textbf{(2) DAgger \citep{ross2011reduction}} 

We use the same state-based teacher as BC and iteratively collect transitions under the student policy, label them with teacher actions at the visited states, aggregate the dataset, and train the student via supervised imitation.

\textbf{(3) COSIL \citep{nguyen2022leveraging}} 
We train the student with RL augmented by a soft imitation term that encourages agreement with a privileged teacher. Following \citet{messikommer2025student}, we use the PPO variant with a fixed target divergence of $0.5$.

\textbf{(4) SITT \citep{messikommer2025student}}
We jointly train the teacher with RL and student with BC and penalize the teacher when its actions are difficult to imitate. The trade-off parameter is selected via grid search.

We use Proximal Policy Optimization (PPO) \citep{schulman2017proximal} for all RL-based methods. All methods use comparable network capacity, matched training budgets, and identical PPO hyperparameters where applicable. Each method is trained with five random seeds and evaluated on 100 environment seeds per run; we report averages across seeds. Full implementation details are provided in Appendix~\ref{appx:implementation}.
The implementation is available at \url{https://github.com/meraccos/RAIL}.
\subsection{\texorpdfstring{Experimental results}{Experimental results}}
\label{sec:results}

\begin{figure}[b]
\centering
\includegraphics[width=1\textwidth]{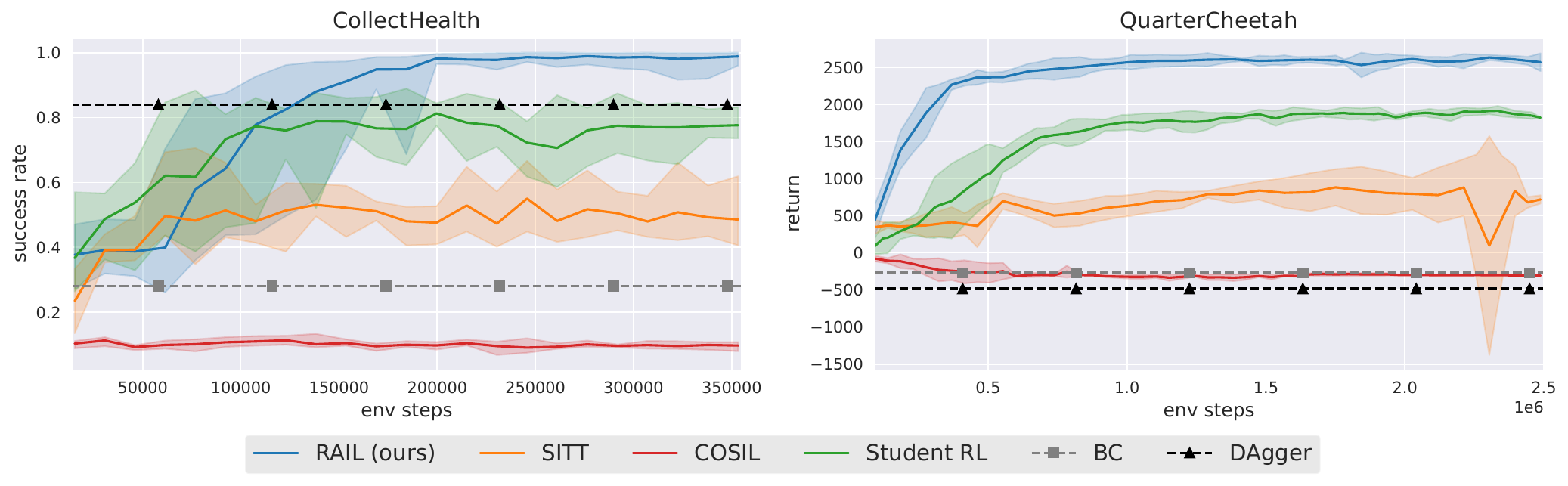}
\caption{Student evaluation curves for CollectHealth and QuarterCheetah. CollectHealth reports success rate, while QuarterCheetah reports episodic return. BC and DAgger are shown as horizontal lines corresponding to their final student performance.
}
\label{fig:curves}
\end{figure}

We report teacher performance ($T$), student performance after transfer ($S$), and the imitation gap ($\Delta = T-S$). For Corners, TunnelVision, and CollectHealth, performance is measured by success rate. For QuarterCheetah, performance is measured by episodic return. Higher $S$ indicates stronger transfer, while lower $\Delta$ indicates more reproducible teacher behavior. Results are summarized in Table~\ref{tab:baselines}, and evaluation curves for the higher-dimensional environments are shown in Figure~\ref{fig:curves}.

\begin{table}[t]
\centering
\tiny
\setlength{\tabcolsep}{2pt}
\renewcommand{\arraystretch}{1.55}
\caption{Teacher performance ($T$), student performance after transfer ($S$), and imitation gap ($\Delta = T-S$). Performance is measured by success rate for Corners, TunnelVision, and CollectHealth, and episodic return for QuarterCheetah. Success rates show the mean with upper and lower semideviations; QuarterCheetah returns show the mean $\pm$ standard deviation. Higher $S$ and lower $\Delta$ are better.}
\label{tab:baselines}

\resizebox{\textwidth}{!}{%
\begin{tabular}{lcccccccccccc}
\toprule
& \multicolumn{3}{c}{\textbf{Corners}} 
& \multicolumn{3}{c}{\textbf{TunnelVision}} 
& \multicolumn{3}{c}{\textbf{CollectHealth}} 
& \multicolumn{3}{c}{\textbf{QuarterCheetah}} \\
\cmidrule(lr){2-4} \cmidrule(lr){5-7} \cmidrule(lr){8-10} \cmidrule(lr){11-13}
Method 
& $T$ & $S$ $\uparrow$ & $\Delta\downarrow$
& $T$ & $S$ $\uparrow$ & $\Delta\downarrow$
& $T$ & $S$ $\uparrow$ & $\Delta\downarrow$
& $T$ & $S$ $\uparrow$ & $\Delta\downarrow$ \\
\midrule
BC             & $1.00^{+0.00}_{-0.00}$ & $0.24^{+0.03}_{-0.02}$ & 0.76 & $1.00^{+0.00}_{-0.00}$ & $0.12^{+0.03}_{-0.03}$ & 0.88 & $1.00^{+0.00}_{-0.00}$ & $0.28^{+0.11}_{-0.09}$ & 0.72 & 2186.83 $\pm$ 82.47 & -268.17 $\pm$ 29.02 & 2455.00 \\[2pt]
DAgger         & $1.00^{+0.00}_{-0.00}$ & $0.38^{+0.15}_{-0.11}$ & 0.62 & $1.00^{+0.00}_{-0.00}$ & $0.15^{+0.02}_{-0.02}$ & 0.85 & $1.00^{+0.00}_{-0.00}$ & $0.84^{+0.02}_{-0.01}$ & 0.16 & 2186.83 $\pm$ 82.47 & -484.25 $\pm$ 23.14 & 2671.08 \\[2pt]
COSIL          & $1.00^{+0.00}_{-0.00}$ & $0.28^{+0.09}_{-0.06}$ & 0.72 & $1.00^{+0.00}_{-0.00}$ & $0.00^{+0.00}_{-0.00}$ & 1.00 & $1.00^{+0.00}_{-0.00}$ & $0.08^{+0.01}_{-0.01}$ & 0.92 & 2186.83 $\pm$ 82.47 & -308.04 $\pm$ 14.46 & 2494.87 \\[2pt]
SITT           & $0.93^{+0.06}_{-0.13}$ & $0.37^{+0.10}_{-0.11}$ & 0.56 & $0.95^{+0.03}_{-0.03}$ & $0.74^{+0.10}_{-0.16}$ & 0.21 & $0.93^{+0.04}_{-0.04}$ & $0.49^{+0.03}_{-0.02}$ & 0.44 & 962.67 $\pm$ 256.17 & 721.12 $\pm$ 239.63 & 241.55 \\[2pt]
RAIL (ours)    & $1.00^{+0.00}_{-0.00}$ & $\mathbf{1.00}^{+0.00}_{-0.00}$ & \textbf{0.00} & $1.00^{+0.00}_{-0.00}$ & $\mathbf{1.00}^{+0.00}_{-0.00}$ & \textbf{0.00} & $0.99^{+0.00}_{-0.01}$ & $\mathbf{0.98}^{+0.01}_{-0.02}$ & \textbf{0.01} & 2627.67 $\pm$ 33.83 & \textbf{2510.24} $\pm$ 72.05 & 117.43 \\[2pt]
\midrule
RAIL w/o stability & $1.00^{+0.00}_{-0.00}$ & $1.00^{+0.00}_{-0.00}$ & 0.00 & $0.55^{+0.29}_{-0.31}$ & $0.53^{+0.16}_{-0.17}$ & 0.02 & $0.97^{+0.02}_{-0.03}$ & $0.95^{+0.04}_{-0.06}$ & 0.02 & 2352.79 $\pm$ 180.64 & 2343.32 $\pm$ 247.17 & \textbf{9.47} \\[2pt]
RAIL w/o alignment & $0.98^{+0.02}_{-0.04}$ & $0.98^{+0.02}_{-0.04}$ & 0.00 & $0.96^{+0.04}_{-0.07}$ & $0.66^{+0.14}_{-0.22}$ & 0.30 & $0.92^{+0.04}_{-0.06}$ & $0.78^{+0.10}_{-0.12}$ & 0.14 & 2634.80 $\pm$ 75.38 & 2476.86 $\pm$ 116.82 & 157.94 \\[2pt]
RAIL end-to-end    & $1.00^{+0.00}_{-0.00}$ & $0.28^{+0.02}_{-0.02}$ & 0.72 & $0.99^{+0.00}_{-0.00}$ & $0.18^{+0.21}_{-0.11}$ & 0.81 & $0.93^{+0.01}_{-0.01}$ & $0.71^{+0.14}_{-0.11}$ & 0.22 & 2283.19 $\pm$ 91.35 & 1898.96 $\pm$ 569.92 & 384.23 \\[2pt]
RAIL sep. encoders & $0.91^{+0.07}_{-0.09}$ & $0.89^{+0.08}_{-0.10}$ & 0.02 & $0.95^{+0.04}_{-0.04}$ & $0.60^{+0.09}_{-0.11}$ & 0.35 & $0.98^{+0.01}_{-0.01}$ & $0.98^{+0.01}_{-0.02}$ & 0.00 & 1813.80 $\pm$ 733.15 & 1690.38 $\pm$ 917.19 & 123.42 \\[2pt]
\midrule
Student RL     & --   & $1.00^{+0.00}_{-0.00}$ & --   & --   & $1.00^{+0.00}_{-0.00}$ & --   & --   & $0.78^{+0.09}_{-0.08}$ & --   & -- & 1824.27 $\pm$ 30.35 & -- \\[2pt]
\bottomrule
\end{tabular}
}
\end{table}

\textbf{Baseline results.} 
BC performs poorly throughout. When the teacher relies on privileged information, identical student observations can correspond to contradictory teacher actions and BC cannot match these from student observations alone.
DAgger improves over BC by mitigating covariate shift through student rollouts, but querying the same privileged teacher leaves this information mismatch unresolved.
COSIL also performs poorly: its auxiliary imitation objective can conflict with reward maximization under student observations by encouraging actions that are suboptimal under student observations.
SITT is the strongest prior baseline, indicating the benefit of changing the teacher's behavior to make it more imitable. However, reward-level regularization remains insufficient when the teacher's private information induces qualitatively different strategies. The student's gains also come with degraded teacher performance, which reflects an explicit trade-off between task optimality and imitability. 
Student RL can solve Corners and TunnelVision better than all other baselines, showing that these approaches can actually impede student performance.

To test whether the failure of the baselines arises from information mismatch, as we argue above, we repeat the training protocol but with matched information content: students receive teacher observations, or unmasked images in QuarterCheetah. In this experiment, all baselines match or approach teacher performance across environments, which supports our explanation (Appendix~\ref{appx:matched-observations}).

\textbf{RAIL results.}
RAIL consistently achieves the highest student performance and the smallest imitation gap across all environments. In Corners and TunnelVision, it attains perfect transfer ($S{=}1.00$) with zero gap, fully matching teacher performance. In CollectHealth, it reaches $S{=}0.98$ with a negligible gap of $0.01$ and substantially outperforms all baselines.
Compared to Student RL, RAIL yields higher performance in CollectHealth ($0.98$ vs.\ $0.78$) and QuarterCheetah (2510.24 vs.\ 1824.27), showing that improved representations and decoupled representation learning can translate into substantially better performance in high-dimensional observation spaces, which is consistent with prior findings in RL~\citep{laskin2020curl, stooke2021decoupling}. 
Notably, RAIL maintains strong teacher performance while achieving strong transfer, unlike SITT, which reduces the gap at the cost of degraded teacher returns. This likely reflects the difficulty of balancing task reward and imitation constraints, together with sensitivity to the reward trade-off.
Figure~\ref{fig:curves} further shows that RAIL reaches strong student performance faster than the online baselines in both high-dimensional environments.

\textbf{Representation analysis.} To assess what information can be recovered from the learned teacher embeddings, we train two-layer prediction networks on frozen embeddings in the Corners environment. Across ten seeds, prediction of the privileged target corner from the initial state achieves $25.4 \pm 1.4\%$ accuracy, close to the $25\%$ chance level, whereas prediction of the agent's coordinates, which are available to both agents, achieves $100.0 \pm 0.0\%$. This provides evidence that the teacher embeddings retain shared state information while suppressing information private to the teacher.

\subsection{Ablation study}
\label{sec:ablation}
The ablations in Table~\ref{tab:baselines} (lower half) isolate the contribution of each component.
Removing the alignment objective (\textbf{RAIL w/o alignment}) significantly reduces transfer performance, particularly in TunnelVision and CollectHealth. Although the contrastive objective encourages agreement, it does not enforce sufficient embedding similarity for direct policy transfer, and small mismatches in the embeddings translate into large imitation gaps.

Removing the stability objective (\textbf{RAIL w/o stability}) has little effect in Corners and CollectHealth, but reduces both teacher and student performance in TunnelVision and QuarterCheetah. The imitation gap nevertheless remains small across all environments, suggesting that stability primarily improves the teacher's training, while alignment helps keep student behavior close to the teacher's.

Allowing gradients from the policy to update the encoder (\textbf{RAIL end-to-end}) leads to large imitation gaps across all tasks despite strong teacher performance. This confirms that RL updates without gradient blocking drive the representation to encode teacher-specific information, resulting in policies that are no longer compatible with student embeddings.

Replacing the shared encoder with separate encoders (\textbf{RAIL sep. encoders}) reduces student performance and increases variability in Corners, TunnelVision, and QuarterCheetah, with little effect in CollectHealth. Consistent with prior work~\citep{poklukar2022geometric}, this suggests that parameter sharing stabilizes the learning of common embeddings.

Finally, to assess RAIL's sensitivity to the contrastive batch size, we evaluate batch sizes from 128 to 1024 while holding the remaining training setup fixed. Performance remains high across this range, with larger batches typically improving success rates and reducing variability (Appendix~\ref{appx:batch-size}).

\subsection{Zero-shot transfer to new tasks}
\label{sec:transfer}
We evaluate whether the learned representations support transferring new behaviors from the teacher agent without collecting additional student data. We modify Corners, TunnelVision and CollectHealth to support two tasks that require different control strategies while sharing the same observation manifold. The task changes involve switching from corner to side targets in Corners, adding diagonal actions in TunnelVision, and switching the rewarded object class in CollectHealth.

We first train RAIL on an initial task, freeze the encoder networks, and then train a new policy from scratch using only teacher observations on the second task. The resulting policy is transferred zero-shot by pairing it with the frozen student encoder. For comparison, we also train complete RAIL, including its encoders and policy, and Student RL from scratch on the second task.

Across all environments, policies trained on the frozen RAIL representation match or surpass the performance of Student RL, despite using no student observations during adaptation. This demonstrates that the learned latent space captures rich environment structure shared across observations rather than features specific to a single objective. The transferred policies also perform on par with RAIL trained on the second task directly, indicating that the representation is sufficiently expressive to support new policies without modification. Detailed descriptions of the modified environments and complete experimental results are provided in Appendix~\ref{appx:transfer}.

\section{Conclusions}
\label{sec:conclusion}
We introduced RAIL, a representation-level approach to teacher--student imitation under asymmetric observations. Rather than training a privileged teacher first and correcting the resulting mismatch afterward, RAIL constrains the teacher to act on representations that are also shared with the student. 
Empirically, RAIL achieves a near-zero imitation gap and outperforms strong baselines across environments. We further show that the learned representation can be reused for new tasks in the same environment through teacher-only policy training.

\subsection*{Limitations}

RAIL assumes access to aligned teacher--student observations generated from the same underlying environment states, which is standard in teacher--student imitation learning but may not hold when observations are collected independently or only partially overlap temporally. Moreover, while our experiments show that gradient blocking and stability regularization substantially reduce the imitation gap, we do not provide formal guarantees that the learned policy is fully realizable under the student observation space.
The zero-shot study evaluates task transfer within the same environments and observation spaces; the conclusions are therefore specific to this setting.

\begin{ack}
This work was partially supported by the Wallenberg AI, Autonomous Systems and Software Program (WASP) funded by the Knut and Alice Wallenberg Foundation.
The authors declare no competing interests.
\end{ack}

\printbibliography

@inproceedings{betser2026infonce,
  title={InfoNCE Induces Gaussian Distribution},
  author={Betser, Roy and Gofer, Eyal and Levi, Meir Yossef and Gilboa, Guy},
  booktitle={International Conference on Learning Representations},
  year={2026},
  note={Published as a conference paper}
}

@inproceedings{chen2019learning,
  author={Chen, Dian and Zhou, Brady and Koltun, Vladlen and Kr{\"a}henb{\"u}hl, Philipp},
  title     = {Learning by Cheating},
  booktitle = {Conference on Robot Learning},
  year      = {2019}
}

@inproceedings{correia2022multiview,
  author={Correia, André and Alexandre, Luís A.},
  booktitle={IEEE International Conference on Robotic Computing}, 
  title={Multi-View Contrastive Learning from Demonstrations}, 
  year={2022},
}

@inproceedings{giammarino2025visually,
  title     = {Visually Robust Adversarial Imitation Learning from Videos with Contrastive Learning},
  author    = {Giammarino, Vittorio and Queeney, James and Paschalidis, Ioannis Ch.},
  booktitle = {International Conference on Robotics and Automation},
  year = {2025},
}

@online{jesse2018generalization,
  author    = {Jesse Farebrother and
               Marlos C. Machado and
               Michael Bowling},
  title     = {Generalization and Regularization in {DQN}},
  year = {2018},
  eprint = {1810.00123},
  eprinttype  = {arxiv},
}

@online{kim2025distilling,
  title={Distilling Realizable Students from Unrealizable Teachers},
  author={Kim, Yujin and Chin, Nathaniel and Vasudev, Arnav and Choudhury, Sanjiban},
  year={2025},
  eprint = {2505.09546},
  eprinttype  = {arxiv},
}

@inproceedings{laskin2020curl,
  title={CURL: Contrastive Unsupervised Representations for Reinforcement Learning},
  author={Laskin, Michael and Srinivas, Aravind and Abbeel, Pieter},
  booktitle={International Conference on Machine Learning},
  year={2020}
}

@inproceedings{messikommer2025student,
  author    = {Nico Messikommer and Jiaxu Xing and Elie Aljalbout and Davide Scaramuzza},
  title     = {Student-Informed Teacher Training},
  booktitle = {International Conference on Learning Representations},
  year      = {2025}
}

@online{messikommer2026approximate,
      title={Approximate Imitation Learning for Event-based Quadrotor Flight in Cluttered Environments}, 
      author={Nico Messikommer and Jiaxu Xing and Leonard Bauersfeld and Marco Cannici and Elie Aljalbout and Davide Scaramuzza},
      year={2026},
  eprint = {2603.07578},
  eprinttype  = {arxiv},
}

@inproceedings{monaci2022dipcan,
    AUTHOR    = {Gianluca Monaci AND Michel Aractingi AND Tomi Silander},
    TITLE     = {{DiPCAN: Distilling Privileged Information for Crowd-Aware Navigation}},
    BOOKTITLE = {Robotics: Science and Systems},
    YEAR      = {2022},
}

@inproceedings{mosbach2024grasp,
  title={Grasp Anything: Combining Teacher-Augmented Policy Gradient Learning with Instance Segmentation to Grasp Arbitrary Objects},
  author={Malte Mosbach and Sven Behnke},
  booktitle={IEEE International Conference on Robotics and Automation},
  year={2024},
}

@inproceedings{nguyen2022leveraging,
  title     = {Leveraging Fully Observable Policies for Learning under Partial Observability},
  author    = {Nguyen, Hai and Baisero, Andrea and Wang, Dian and Amato, Christopher and Platt, Robert},
  booktitle = {Conference on Robot Learning},
  year      = {2022},
}

@online{oord2018representation,
  title={Representation Learning with Contrastive Predictive Coding},
  author={Oord, Aaron van den and Li, Yazhe and Vinyals, Oriol},
  year={2018},
  eprint = {1807.03748},
  eprinttype  = {arxiv},
}

@inproceedings{pomerleau1989alvinn,
  title={ALVINN: An autonomous land vehicle in a neural network},
  author={Pomerleau, Dean A.},
  booktitle={Advances in Neural Information Processing Systems},
  year={1989}
}

@inproceedings{radford2021learning,
  title={Learning Transferable Visual Models From Natural Language Supervision},
  author={Radford, Alec and Kim, Jong Wook and Hallacy, Chris and Ramesh, Aditya and Goh, Gabriel and Agarwal, Sandhini and Sastry, Girish and Askell, Amanda and Mishkin, Pamela and Clark, Jack and Krueger, Gretchen and Sutskever, Ilya},
  booktitle={International Conference on Machine Learning},
  year={2021},
}

@inproceedings{ross2011reduction,
  author    = {St{\'e}phane Ross and Geoffrey J. Gordon and J. Andrew Bagnell},
  title     = {A Reduction of Imitation Learning and Structured Prediction to No-Regret Online Learning},
  booktitle = {International Conference on Artificial Intelligence and Statistics},
  year      = {2011}
}

@online{schulman2017proximal,
  author    = {John Schulman and Filip Wolski and Prafulla Dhariwal and Alec Radford and Oleg Klimov},
  title     = {Proximal Policy Optimization Algorithms},
  year      = {2017},
  eprint = {1707.06347},
  eprinttype  = {arXiv},
}

@inproceedings{sermanet2018time,
  title     = {Time-Contrastive Networks: Self-Supervised Learning from Video},
  author    = {Sermanet, Pierre and Lynch, Corey and Chebotar, Yevgen and Hsu, Jasmine and Jang, Eric and Schaal, Stefan and Levine, Sergey},
  booktitle = {IEEE International Conference on Robotics and Automation},
  year      = {2018},
}

@inproceedings{shon2006learning,
  title     = {Learning Shared Latent Structure for Image Synthesis and Robotic Imitation},
  author    = {Shon, Aaron P. and Grochow, Keith and Hertzmann, Aaron and Rao, Rajesh P. N.},
  booktitle = {Advances in Neural Information Processing Systems},
  year      = {2005},
}

@inproceedings{song2023learning,
  author={Song, Yunlong and Shi, Kexin and Penicka, Robert and Scaramuzza, Davide},
  title     = {Learning Perception-Aware Agile Flight in Cluttered Environments},
  booktitle = {IEEE International Conference on Robotics and Automation},
  year      = {2023},
}

@inproceedings{song2026to,
 author = {Song, Yuda and Rohatgi, Dhruv and Singh, Aarti and Bagnell, J.},
 booktitle = {Advances in Neural Information Processing Systems},
 doi = {10.52202/085713-2996},
 pages = {89559--89606},
 title = {To Distill or Decide? Understanding the Algorithmic Trade-off in Partially Observable RL},
 volume = {38, Main Conference},
 year = {2025}
}

@inproceedings{stadie2017third,
  title     = {Third-Person Imitation Learning},
  author    = {Stadie, Bradly C. and Abbeel, Pieter and Sutskever, Ilya},
  booktitle = {International Conference on Learning Representations},
  year      = {2017}
}

@inproceedings{stooke2021decoupling,
  title={Decoupling Representation Learning from Reinforcement Learning},
  author={Stooke, Adam and Lee, Kimin and Abbeel, Pieter and Laskin, Michael},
  booktitle={International Conference on Machine Learning},
  year={2021}
}

@inproceedings{todorov2012mujoco,
  title={MuJoCo: A physics engine for model-based control},
  author={Todorov, Emanuel and Erez, Tom and Tassa, Yuval},
  booktitle={International Conference on Intelligent Robots and Systems},
  year={2012},
}

@inproceedings{towers2025gymnasium,
  title = {Gymnasium: A Standard Interface for Reinforcement Learning Environments},
  author = {
    Towers, Mark and
    Kwiatkowski, Ariel and
    Terry, Jordan K. and
    Balis, John U. and
    De Cola, Gianluca and
    Deleu, Tristan and
    Goul{\~a}o, Manuel and
    Kallinteris, Andreas and
    Krimmel, Markus and
    KG, Arjun and
    Perez-Vicente, Rodrigo and
    Pierr{\'e}, Andrea and
    Schulhoff, Sander and
    Tai, Jun Jet and
    Tan, Hannah and
    Younis, Omar G.
  },
  booktitle = {Advances in Neural Information Processing Systems},
  volume = {38},
  year = {2025},
  doi = {10.52202/085713-4916},
}

@inproceedings{vonkugelgen2021contentstyle,
  author={von K{\"u}gelgen, Julius and Sharma, Yash and Gresele, Luigi and Brendel, Wieland and Sch{\"o}lkopf, Bernhard and Besserve, Michel and Locatello, Francesco},
  title     = {Self-Supervised Learning with Data Augmentations Provably Isolates Content from Style},
  booktitle = {Advances in Neural Information Processing Systems},
  year      = {2021}
}

@inproceedings{
walsman2023impossibly,
title={Impossibly Good Experts and How to Follow Them},
author={Aaron Walsman and Muru Zhang and Sanjiban Choudhury and Dieter Fox and Ali Farhadi},
booktitle={International Conference on Learning Representations},
year={2023},
}

@inproceedings{weihs2021bridging,
title={Bridging the Imitation Gap by Adaptive Insubordination},
author={Luca Weihs and Unnat Jain and Iou-Jen Liu and Jordi Salvador and Svetlana Lazebnik and Aniruddha Kembhavi and Alex Schwing},
booktitle={Advances in Neural Information Processing Systems},
year={2021},
}

@inproceedings{xing2024bootstrapping,
  title     = {Bootstrapping Reinforcement Learning with Imitation for Vision-Based Agile Flight},
  author    = {Xing, Jiaxu and Romero, Angel and Bauersfeld, Leonard and Scaramuzza, Davide},
  booktitle = {Conference on Robot Learning},
  year      = {2024},
}

@inproceedings{xu2020error,
  title     = {Error Bounds of Imitating Policies and Environments},
  author    = {Xu, Tian and Li, Ziniu and Yu, Yang},
  booktitle = {Advances in Neural Information Processing Systems},
  year      = {2020},
}

@inproceedings{zhang2021roach,
  title = {End-to-End Urban Driving by Imitating a Reinforcement Learning Coach},
  booktitle = {IEEE/CVF International Conference on Computer Vision},
  author = {Zhang, Zhejun and Liniger, Alexander and Dai, Dengxin and Yu, Fisher and Van Gool, Luc},
  year = {2021},
}

@online{maxime2023mini,
      title={Minigrid \& Miniworld: Modular \& Customizable Reinforcement Learning Environments for Goal-Oriented Tasks}, 
      author    = {Chevalier-Boisvert, Maxime and Dai, Bolun and Towers, Mark and de Lazcano, Rodrigo and Willems, Lucas and Lahlou, Salem and Pal, Suman and Castro, Pablo Samuel and Terry, Jordan},
      year={2023},
      eprint={2306.13831},
      eprinttype  = {arxiv},
}

@inproceedings{paszke2019pytorch,
  title     = {PyTorch: An Imperative Style, High-Performance Deep Learning Library},
  author    = {Paszke, Adam and Gross, Sam and Massa, Francisco and Lerer, Adam and Bradbury, James and Chanan, Gregory and Killeen, Trevor and Lin, Zeming and Gimelshein, Natalia and Antiga, Luca and others},
  booktitle = {Advances in Neural Information Processing Systems},
  year      = {2019}
}

@article{raffin2021stable,
  title   = {Stable-Baselines3: Reliable Reinforcement Learning Implementations},
  author  = {Raffin, Antonin and Hill, Ashley and Gleave, Adam and Kanervisto, Anssi and Ernestus, Maximilian and Dormann, Noah},
  journal = {Journal of Machine Learning Research},
  volume  = {22},
  number  = {268},
  pages   = {1--8},
  year    = {2021},
}

@inproceedings{poklukar2022geometric,
  title = {Geometric Multimodal Contrastive Representation Learning},
  author = {Poklukar, Petra and Vasco, Miguel and Yin, Hang and Melo, Francisco S. and Paiva, Ana and Kragic, Danica},
  booktitle = {International Conference on Machine Learning},
  year = {2022},
}

\clearpage

\appendix

\section{Environments}
\label{appx:envs}

\textbf{Corners} is a grid-world navigation task on an \(n \times n\) grid (\(n{=}9\)), where the agent starts at the center and must reach a target corner sampled uniformly at the beginning of each episode. The action space consists of 8 discrete movements (cardinal and diagonal), with positions clipped to remain within the grid. Observations encode the agent position \((x,y)\) as one-hot vectors of size \(n\) each, and the previous action as an 8-dimensional one-hot vector over \(\{0,\dots,7\}\). The teacher additionally observes the target corner as a 4-dimensional one-hot vector, while the student does not, yielding \(\dim(o^T)=2n+12\) and \(\dim(o^S)=2n+8\). This private goal information induces an imitation gap, as the teacher can move directly to the target while the student must infer it through exploration. Episodes terminate upon reaching the target and are truncated after 60 steps. The reward function is \(r_t = \mathbf{1}\{\text{goal reached}\} - 0.01\).

\textbf{TunnelVision} is a grid-world navigation task on an \(n \times n\) grid (\(n{=}8\)) surrounded by a one-cell obstacle border, where the agent starts at \((1,1)\) and must reach the fixed goal at \((n-2,n-2)\). Obstacles are sampled randomly at reset with probability \(0.5\), subject to constraints that remove diagonal obstacle patterns and preserve a valid path from start to goal. The agent can move or rotate in the four compass directions, yielding 8 discrete actions. 
Observations contain the local \(3\times3\) occupancy grid, 4-dimensional one-hot-coded orientation, normalized previous action, and normalized position. The teacher observes all 9 cells in the local neighborhood, while the student observes only the 3 cells in front of its current orientation, with unseen cells masked by a constant; both observations have dimension \(\dim(o^T)=\dim(o^S)=16\). 
This asymmetry in information induces an imitation gap, as the teacher can solve the task without turning while the student must reorient to reveal hidden obstacles. 
Episodes terminate when the agent reaches the goal or collides with an obstacle/border and are truncated after 80 steps. The reward function is \(r_t = 1\times\mathbf{1}\{\text{goal reached}\} - 1\times\mathbf{1}\{\text{collision}\} - 0.01\times\mathbf{1}\{d_{t+1}\ge d_t\}\), where \(d_t\) is the Euclidean distance to the goal before the action.

\textbf{CollectHealth} is an episodic 3D navigation task based on MiniWorld~\citep{maxime2023mini}, where the agent operates in a \(15\text{ m}\times15\text{ m}\) room and must collect \(5\) randomly placed health boxes. The action space has 10 discrete actions: movement in 8 uniformly spaced relative directions with step size \(0.5\text{ m}\), and two camera rotations of \(\pm 30^\circ\). The teacher observes the relative positions of all remaining boxes in the agent frame, sorted by distance and padded with \(-1\), together with the normalized agent position and previous action, giving \(\dim(o^T)=13\). 
The student observes a stack of 5 RGB first-person images, each resized to \(84\times84\) and normalized to \([-1,1]\), giving \(\dim(o^S)=5\times3\times84\times84\).
This extra target information induces an imitation gap, as the teacher can move directly toward unseen boxes while the student must actively explore to locate them.
A box is collected when its distance to the agent is below \(1.0\text{m}\). Episodes terminate when all boxes are collected or when the agent collides with a wall, and are truncated after 200 steps. The reward function is \(r_t=10\,\times\mathbf{1}\{\text{all boxes collected}\}+1\times\mathbf{1}\{\text{one box collected}\}-5\times\mathbf{1}\{\text{wall collision}\}-0.05\). Success is defined as the fraction of collected boxes, with terminated episodes assigned zero success.

\textbf{QuarterCheetah} is a continuous-control task based on HalfCheetah~\citep{towers2025gymnasium}, simulated in MuJoCo~\citep{todorov2012mujoco}. The agent must maximize forward velocity using a 6-dimensional continuous action space, \(\mathcal A \subset \mathbb R^6\). 
The teacher observes the full 17-dimensional proprioceptive state (joint positions and velocities), while the student observes only the right half of a stack of 4 RGB renders, each resized to \(84\times84\) and normalized to \([-1,1]\), yielding \(\dim(o^T)=17\) and \(\dim(o^S)=4\times3\times84\times84\).
This asymmetry in the observations induces an imitation gap, as the teacher can act directly on full state information while the student must find a working policy while seeing only the right half of the visual input.
Episodes follow the default HalfCheetah horizon (1000 steps). The reward function is the standard MuJoCo HalfCheetah reward: \(r_t = v_t - 0.1 \|a_t\|^2\), where \(v_t\) is forward velocity and \(a_t\) is the action.

\newpage

\section{Zero-shot transfer to new tasks}
\label{appx:transfer}
This appendix provides additional details for the zero-shot transfer experiment described in Section~\ref{sec:transfer}. The goal of this experiment is to evaluate whether the representation learned by RAIL can be reused to learn new behaviors through teacher-only training and transferred directly to the student agent.

\subsection{Modified environments}
To evaluate the transfer capabilities of RAIL, we construct modified versions of Corners, TunnelVision, and CollectHealth environments (\cf Appendix~\ref{appx:envs}) that support two distinct tasks within the same observation space. The tasks require different behaviors while preserving the same underlying observation manifolds.

\textbf{CornersPlus.}
We modify Corners by introducing an additional set of target locations at the midpoints of each wall. At the start of each episode, both a corner target and a side target are sampled uniformly and revealed only to the teacher. In Task 1, the agent must reach the corner target while avoiding side targets; in Task 2, it must reach the side target while avoiding corner targets. 
To ensure that the required behaviors differ substantially, visiting a side target in Task 1 or a corner target in Task 2 immediately terminates the episode as failure. 
The modified reward function is \(r_t = 1\times\mathbf{1}\{\text{goal reached}\} - 0.1\times\,\mathbf{1}\{\text{termination}\} - 0.01\).

\textbf{TunnelVisionPlus.}
We modify TunnelVision by changing the action space while keeping observations unchanged. In Task 1, as in the original TunnelVision, the agent operates with 8 actions (four movements and four rotations). In Task 2, four diagonal movement actions are added, increasing the action space to 12. The reward and termination conditions remain identical to TunnelVision.

\textbf{CollectHealthPlus.}
We modify CollectHealth by adding an equal number of secondary objects into the environment (rubber ducks) and changing the rewarded object class. In Task 1, medkits are rewarded and ducks penalized; in Task 2, ducks are rewarded and medkits penalized. Both medkits and ducks are randomly placed with a minimum separation of \(3.0\)m, and the agent is initialized at least \(3.0\)m away from all objects. 
Unlike the original CollectHealth, we place only two objects of each type to reduce scene density and avoid trivial collection strategies.
The modified reward function is \(r_t = 10\times\mathbf{1}\{\text{all target objects collected}\} + 1\times\mathbf{1}\{\text{target collected}\} - 1\times\mathbf{1}\{\text{incorrect object collected}\lor\text{wall collision}\} - 0.01\).

\subsection{Transfer results}
Table~\ref{tab:transfer} shows that the representations learned by RAIL are not tied to a single policy or reward function, but instead capture rich environment details that support new behaviors without additional student-side training. In all three environments, training a new policy on top of a frozen representation learned from Task 1 yields nearly identical student performance to training RAIL from scratch on Task 2. This is notable because, after freezing the encoders, training the new policy is performed entirely using teacher observations. Nevertheless, the resulting policies remain directly executable from student observations, which indicate that the learned latent space preserves the shared environment structure required for learning new behaviors while maintaining teacher--student alignment.

\begin{table}[h]
\caption{Student success rates on the new task. Task 2 denotes direct training on the second task from scratch. Task $1\rightarrow2$ denotes transfer after first learning the representation on Task 1, freezing both encoders, training a new Task 2 policy using only teacher observations, and deploying zero-shot to the student. Matching Task $1\rightarrow2$ to Task 2 indicates successful representation reuse. Rates show the mean with upper and lower semideviations across five runs.}
\centering
\small
\begin{tabular}{lcccccc}
\toprule
& \multicolumn{2}{c}{\textbf{CornersPlus}} & \multicolumn{2}{c}{\textbf{TunnelVisionPlus}} & \multicolumn{2}{c}{\textbf{CollectHealthPlus}} \\
\cmidrule(lr){2-3} \cmidrule(lr){4-5} \cmidrule(lr){6-7}
Method & Task 2 & Task 1$\rightarrow$2 & Task 2 & Task 1$\rightarrow$2 & Task 2 & Task 1$\rightarrow$2 \\
\midrule
RAIL (student) & $1.0^{+0.00}_{-0.00}$ & $1.0^{+0.00}_{-0.00}$ & $1.0^{+0.00}_{-0.00}$ & $1.0^{+0.00}_{-0.00}$ & $0.84^{+0.07}_{-0.07}$ & $0.83^{+0.06}_{-0.06}$ \\[2pt]
Student RL     & $1.0^{+0.00}_{-0.00}$ & -- & $1.0^{+0.00}_{-0.00}$ & -- & $0.81^{+0.11}_{-0.11}$ & -- \\[2pt]
\bottomrule
\end{tabular}
\label{tab:transfer}
\end{table}

\newpage

\section{Implementation details}
\label{appx:implementation}

\subsection{InfoNCE implementation}
\label{appx:infonce}
We use a symmetric InfoNCE objective following the bidirectional formulation of CLIP \citep{radford2021learning}. Given minibatch embeddings \(\{z_T^i,z_S^i\}_{i=1}^N\), we first \(\ell_2\)-normalize both embedding sets and compute similarity logits
\begin{equation}
M_{ij}= s\,(z_T^i)^\top z_S^j,
\end{equation}
where \(s=\exp(\alpha)=1/\tau\) is a learnable logit scale with unconstrained parameter \(\alpha\). We initialize \(s=1/0.05\), optimize \(\alpha\) jointly with all parameters, and clamp \(s \leq 50\) for numerical stability.

The contrastive objective averages teacher-to-student and student-to-teacher cross-entropies:
\begin{equation}
\mathcal{L}_{\mathrm{InfoNCE}}
=
\frac{1}{2}\Big[
\mathrm{CE}(M,y)+\mathrm{CE}(M^\top,y)
\Big],
\end{equation}
where CE is the cross entropy loss and \(y=(1,\dots,N)\) specifies that sample \(i\) is matched to its paired sample \(i\).

\paragraph{Multi-positive variant.}
In environments with small or discrete observation spaces, multiple samples in a minibatch may correspond to identical teacher or student observations. Treating these duplicates as negatives creates false-negative pairs. We therefore replace the one-hot target with a multi-positive mask \(P \in \{0,1\}^{N \times N}\):
\begin{equation}
P_{ij}=\mathbf{1}\!\left[o_T^i=o_T^j \;\lor\; o_S^i=o_S^j \right].
\end{equation}

The teacher-to-student term becomes
\begin{equation}
\mathcal{L}_{T \rightarrow S}
=
-\frac{1}{N}\sum_{i=1}^{N}
\log
\sum_{j=1}^{N}
P_{ij}
\frac{\exp(M_{ij})}{\sum_{k=1}^{N}\exp(M_{ik})},
\end{equation}
and the reverse-direction term is
\begin{equation}
\mathcal{L}_{S \rightarrow T}
=
-\frac{1}{N}\sum_{i=1}^{N}
\log
\sum_{j=1}^{N}
P_{ji}
\frac{\exp(M_{ji})}{\sum_{k=1}^{N}\exp(M_{ki})}.
\end{equation}

The final objective is
\begin{equation}
\mathcal{L}_{\mathrm{MP\text{-}InfoNCE}}
=
\frac{1}{2}
\left(
\mathcal{L}_{T \rightarrow S}
+
\mathcal{L}_{S \rightarrow T}
\right).
\end{equation}

We use this variant only in Corners and TunnelVision, where repeated observations are common.

\subsection{Network architectures}
\label{appx:architectures}

All methods use comparable model capacity. RAIL consists of a teacher and a student encoder, a shared encoder, and a policy network with actor--critic heads.

\paragraph{Vector observations.}
For state-based agents, each observation encoder is a three-layer MLP with hidden width 512 and LeakyReLU activations: $d_{\mathrm{obs}} \rightarrow 512 \rightarrow 512 \rightarrow 512.$

\paragraph{Pixel observations.}
For pixel observations, we use a four-layer CNN with \(3 \times 3\) kernels, 32 channels per layer, and strides \((2,2,1,1)\), followed by a linear layer producing a 512-dimensional feature vector.

\paragraph{Shared encoder.}
Teacher and student features are concatenated with a learned agent-identity embedding of dimension 4, then mapped by a shared MLP: $516 \rightarrow 512 \rightarrow 512 \rightarrow d_z, $
followed by \(\ell_2\)-normalization for contrastive learning.

\paragraph{Policy network.}
We use the standard PPO policy architecture: two shared hidden layers of size 64 with ReLU activations, followed by separate linear actor and critic output heads.

\subsection{Training hyperparameters}
\label{appx:hyperparameters}
We use PPO for all RL-based methods and Adam for both policy and encoder optimization. 
Table~\ref{tab:hyperparams} lists the environment-specific training budgets and hyperparameters.
Unless stated otherwise, all remaining hyperparameters use the default values from PyTorch~\citep{paszke2019pytorch} and Stable-Baselines3~\citep{raffin2021stable}.

For QuarterCheetah, we found that direct Student RL and the privileged teacher training used for BC became unstable with the standard PPO learning rate. 
We therefore reduced the PPO learning rate to \(2\times10^{-5}\) for these settings to prevent training collapse.

All experiments were conducted on a machine with an Intel Core i7-13700KF CPU (16 cores, 24 threads, up to 5.4 GHz), an NVIDIA RTX 4070 GPU with 12 GB VRAM, and 32 GB RAM. Depending on the environment and method, individual runs required approximately 1--7 hours. In QuarterCheetah, RAIL required approximately four hours, compared with seven hours for standard PPO from pixels under the same environment-interaction budget.

The BC datasets are collected from rollouts of the trained state-based teacher and contain 4,096 transitions for Corners and TunnelVision, 20,000 for CollectHealth, and 80,000 for QuarterCheetah. DAgger uses replay buffers of the same respective sizes, populated from student rollouts.

\vspace{5pt}
\begin{table}[h]
\centering
\caption{Training hyperparameters for each environment.}
\label{tab:hyperparams}
\small
\setlength{\tabcolsep}{5pt}
\begin{tabular}{lcccc}
\toprule
 & Corners & TunnelVision & CollectHealth & QuarterCheetah \\
\midrule

training steps & 1.0M & 1.5M & 0.4M & 2.5M \\
parallel environments & 12 & 12 & 12 & 4 \\

\midrule
\multicolumn{5}{l}{\textbf{PPO}} \\

optimizer & Adam & Adam & Adam & Adam \\
learning rate & \(3 \times 10^{-4}\) & \(3 \times 10^{-4}\) & \(3 \times 10^{-4}\) & \(3 \times 10^{-4}\) \\
batch size & 256 & 512 & 256 & 64 \\
rollout steps ($n_{\mathrm{steps}}$) & 256 & 512 & 256 & 512 \\
epochs & 10 & 10 & 10 & 20 \\
entropy coefficient & 0.002 & 0.001 & 0.000 & 0.0004 \\
clip range & 0.1 & 0.1 & 0.1 & 0.1 \\
discount factor ($\gamma$) & 0.99 & 0.99 & 0.99 & 0.98 \\

\midrule
\multicolumn{5}{l}{\textbf{RAIL}} \\

encoder optimizer & Adam & Adam & Adam & Adam \\
encoder learning rate & \(10^{-3}\) & \(10^{-3}\) & \(10^{-3}\) & \(10^{-3}\) \\
embedding dimension ($d_z$) & 64 & 128 & 256 & 256 \\
contrastive batch size & 512 & 512 & 256 & 512 \\
contrastive iterations ($K$) & 1 & 2 & 12 & 8 \\
$\lambda_{\mathrm{con}}$ & 1.0 & 1.0 & 1.0 & 1.0 \\
$\lambda_{\mathrm{align}}$ & 1.0 & 1.0 & 1.0 & 1.0 \\
$\lambda_{\mathrm{stab}}$ & 1.0 & 1.0 & 1.0 & 1.0 \\
initial InfoNCE scale & 20 & 20 & 20 & 20 \\
maximum InfoNCE scale & 50 & 50 & 50 & 50 \\

\bottomrule
\end{tabular}
\end{table}

\clearpage

\section{\texorpdfstring{Additional experiments}{Additional experiments}}
\label{appx:additional}

\subsection{\texorpdfstring{Privileged-information probes}{Privileged-information probes}}
\label{appx:probing}

We use Corners to test whether teacher embeddings retain shared information while suppressing information private to the teacher. The target corner is sampled uniformly from four possibilities and observed only by the teacher, whereas the agent's coordinates are available to both agents. We train RAIL from scratch with ten random seeds and freeze the learned encoders. For each model, we collect probe data from 100 independently initialized episodes with independently sampled target corners.

We train separate two-layer MLP probes on the frozen teacher embeddings to predict the target corner and the agent's coordinates. We evaluate target-corner prediction only at the initial state, where the agent always starts at the grid center. This prevents the probe from exploiting correlations between the agent's position and target corner that arise during a rollout. The coordinate probe uses embeddings from the full rollout buffer to assess whether the representation preserves shared spatial information.

Across ten seeds, target-corner prediction achieves $25.4 \pm 1.4\%$ accuracy, close to the $25\%$ chance level, while coordinate prediction achieves $100.0 \pm 0.0\%$. These results provide evidence that RAIL preserves shared information while suppressing information private to the teacher.

\subsection{\texorpdfstring{Contrastive batch size}{Contrastive batch size}}
\label{appx:batch-size}

The number of negatives in our contrastive objective (Eq. \eqref{eq:contrastive}) depends on the batch size used for representation learning. We experiment with batch sizes of 128, 256, 512, and 1024, keeping the remaining training setup fixed and using five random seeds per setting. Table~\ref{tab:batch-size} reports student performance.

RAIL achieves high student performance throughout this range. In Corners, batch sizes of 512 and 1024 yield perfect success in all five runs, whereas smaller batches have lower mean success and greater variability. TunnelVision and CollectHealth approach saturation at moderate batch sizes, and QuarterCheetah returns remain similar across settings. Thus, larger batches can improve reliability, while RAIL maintains strong performance even with smaller batches.

\begin{table}[htbp]
\centering
\caption{Contrastive batch-size ablation. Student success rates show the mean with upper and lower semideviations over five seeds; QuarterCheetah episodic returns show mean $\pm$ standard deviation.}
\label{tab:batch-size}
\small
\begin{tabular}{rcccc}
\toprule
Batch size & Corners & TunnelVision & CollectHealth & QuarterCheetah \\
\midrule
128  & $0.91^{+0.08}_{-0.16}$ & $0.99^{+0.01}_{-0.01}$ & $0.94^{+0.03}_{-0.04}$ & $2491.33 \pm 154.06$ \\[2pt]
256  & $0.89^{+0.08}_{-0.10}$ & $1.00^{+0.00}_{-0.00}$ & $0.98^{+0.00}_{-0.01}$ & $2523.33 \pm 84.63$ \\[2pt]
512  & $1.00^{+0.00}_{-0.00}$ & $1.00^{+0.00}_{-0.00}$ & $0.99^{+0.00}_{-0.01}$ & $2510.24 \pm 72.05$ \\[2pt]
1024 & $1.00^{+0.00}_{-0.00}$ & $1.00^{+0.00}_{-0.00}$ & $0.99^{+0.01}_{-0.01}$ & $2543.33 \pm 45.76$ \\[2pt]
\bottomrule
\end{tabular}
\end{table}

\clearpage
\subsection{\texorpdfstring{Baseline experiments under matched observations}{Baseline controls under matched observations}}
\label{appx:matched-observations}

To test whether the failure of the baselines arises from information mismatch, we repeat BC, DAgger, COSIL, and SITT with the same training protocols but matched information content. Students receive the teacher's observations in Corners, TunnelVision, and CollectHealth. In QuarterCheetah, students receive stacked, unmasked RGB images while teachers retain state observations.

All four methods match or approach teacher performance across all environments (Table~\ref{tab:matched-observations}). This contrast with Table~\ref{tab:baselines} supports that the information mismatch is a principal contributor to the failure of baselines in our environments.

\begin{table}[htbp]
\centering
\tiny
\setlength{\tabcolsep}{2pt}
\caption{Baseline performance under matched observations. Asterisks denote environments with modified student observations. Teacher ($T$) and student ($S$) success rates show the mean with upper and lower semideviations over five seeds; QuarterCheetah returns show mean $\pm$ standard deviation. The imitation gap is $\Delta=T-S$. Higher $S$ and lower $\Delta$ are better.}
\label{tab:matched-observations}
\resizebox{\textwidth}{!}{%
\begin{tabular}{lcccccccccccc}
\toprule
& \multicolumn{3}{c}{\textbf{Corners}$^{\ast}$}
& \multicolumn{3}{c}{\textbf{TunnelVision}$^{\ast}$}
& \multicolumn{3}{c}{\textbf{CollectHealth}$^{\ast}$}
& \multicolumn{3}{c}{\textbf{QuarterCheetah}$^{\ast}$} \\
\cmidrule(lr){2-4} \cmidrule(lr){5-7} \cmidrule(lr){8-10} \cmidrule(lr){11-13}
Method
& $T$ & $S$ $\uparrow$ & $\Delta\downarrow$
& $T$ & $S$ $\uparrow$ & $\Delta\downarrow$
& $T$ & $S$ $\uparrow$ & $\Delta\downarrow$
& $T$ & $S$ $\uparrow$ & $\Delta\downarrow$ \\
\midrule
BC & $1.00^{+0.00}_{-0.00}$ & $1.00^{+0.00}_{-0.00}$ & 0.00 & $1.00^{+0.00}_{-0.00}$ & $1.00^{+0.00}_{-0.00}$ & 0.00 & $1.00^{+0.00}_{-0.00}$ & $1.00^{+0.00}_{-0.00}$ & 0.00 & 2186.83 $\pm$ 82.47 & 2113.43 $\pm$ 57.39 & 73.40 \\[2pt]
DAgger & $1.00^{+0.00}_{-0.00}$ & $1.00^{+0.00}_{-0.00}$ & 0.00 & $1.00^{+0.00}_{-0.00}$ & $1.00^{+0.00}_{-0.00}$ & 0.00 & $1.00^{+0.00}_{-0.00}$ & $1.00^{+0.00}_{-0.00}$ & 0.00 & 2186.83 $\pm$ 82.47 & 2099.67 $\pm$ 62.27 & 87.16 \\[2pt]
COSIL & $1.00^{+0.00}_{-0.00}$ & $1.00^{+0.00}_{-0.00}$ & 0.00 & $1.00^{+0.00}_{-0.00}$ & $1.00^{+0.00}_{-0.00}$ & 0.00 & $1.00^{+0.00}_{-0.00}$ & $1.00^{+0.00}_{-0.00}$ & 0.00 & 2186.83 $\pm$ 82.47 & 2090.52 $\pm$ 68.50 & 96.31 \\[2pt]
SITT & $1.00^{+0.00}_{-0.00}$ & $1.00^{+0.00}_{-0.00}$ & 0.00 & $1.00^{+0.00}_{-0.00}$ & $1.00^{+0.00}_{-0.00}$ & 0.00 & $1.00^{+0.00}_{-0.00}$ & $1.00^{+0.00}_{-0.00}$ & 0.00 & 1865.42 $\pm$ 86.72 & 1720.43 $\pm$ 74.45 & 144.99 \\[2pt]
\bottomrule
\end{tabular}
}
\end{table}

\clearpage

\end{document}